\documentclass[letterpaper, 10 pt, conference]{ieeeconf}  

\IEEEoverridecommandlockouts                              

\usepackage{cite}
\usepackage{mathtools}
\usepackage[caption=false]{subfig}

\usepackage{multirow}
\usepackage{makecell}

\usepackage{newclude}
\usepackage{booktabs}

\title{\LARGE \bf
Colonoscopy Navigation using End-to-End Deep Visuomotor Control:\\ A User Study}

\author{ Ameya Pore$^{1,2,*}$, Martina Finocchiaro $^{2,3}$, Diego Dall'Alba$^{1}$, Albert Hernansanz$^{2}$, Gastone Ciuti$^{3}$,\\ Alberto Arezzo$^{4}$, Arianna Menciassi$^{3}$, Alicia Casals$^{2}$, Paolo Fiorini$^{1}$
\thanks{This project has received funding from the European Union’s Horizon 2020 research and innovation programme under the Marie Skłodowska-Curie (grant agreement No. 813782 "ATLAS")}
\thanks{* Corresponding author: Ameya Pore (email: ameya.pore@univr.it)}
\thanks{$^{1}$ Department of Computer Science, University of Verona, Italy}
\thanks{$^{2}$ Department of Enginyeria de Sistemas, Automatica i Informatica Industrial, Univiversitat Politecnica de Catalunya, Barcelona, Spain }
\thanks{$^{3}$ The BioRobotics Institute, Scuola Superiore Sant’Anna, Pisa, Italy }
\thanks{$^{4}$ Department of Surgical Sciences, University of Torino, Turin, Italy}
}

\begin{document}

\maketitle
\thispagestyle{empty}
\pagestyle{empty}

\begin{abstract}

Flexible endoscopes for colonoscopy present several limitations due to their inherent complexity, resulting in patient discomfort and lack of intuitiveness for clinicians.
Robotic devices together with autonomous control represent a viable solution to reduce the workload of endoscopists and the training time while improving the overall procedure outcome.
Prior works on autonomous endoscope control use heuristic policies that limit their generalisation to the unstructured and highly deformable colon environment and require frequent human intervention.
This work proposes an image-based control of the endoscope using Deep Reinforcement Learning, called Deep Visuomotor Control (DVC), to exhibit adaptive behaviour in convoluted sections of the colon tract. 
DVC learns a mapping between the endoscopic images and the control signal of the endoscope.
A first user study of 20 expert gastrointestinal endoscopists was carried out to compare their navigation performance with DVC policies using a realistic virtual simulator.
The results indicate that DVC shows equivalent performance on several assessment parameters, being more safer. 
Moreover, a second user study with 20 novice participants was performed to demonstrate easier human supervision compared to a state-of-the-art heuristic control policy.  
Seamless supervision of colonoscopy procedures would enable interventionists to focus on the medical decision rather than on the control problem of the endoscope.
\end{abstract}

\section{INTRODUCTION}

Colonoscopy screening programs remain the gold standard for the diagnosis and treatment of lower-gastric pathologies such as colorectal cancer (CRC), which is the third most common malignancy worldwide \cite{bray2018global}. During a routine diagnostic procedure, a Flexible Endoscope (FE) is firstly inserted from the rectum to the caecum and then retracted slowly to detect possible early-stage CRC lesions. Early-stage detection of CRC can improve the five-year survival rate by 90\% \cite{chen2017toll}. 
FE-based procedures are complex due to non-intuitive mapping between the endoscope tip and the control steering knobs, which requires a long and extensive training process to be mastered \cite{ciuti2020frontiers}.
\begin{figure}[thpb]
	\centering
	\includegraphics[width=0.49\textwidth]{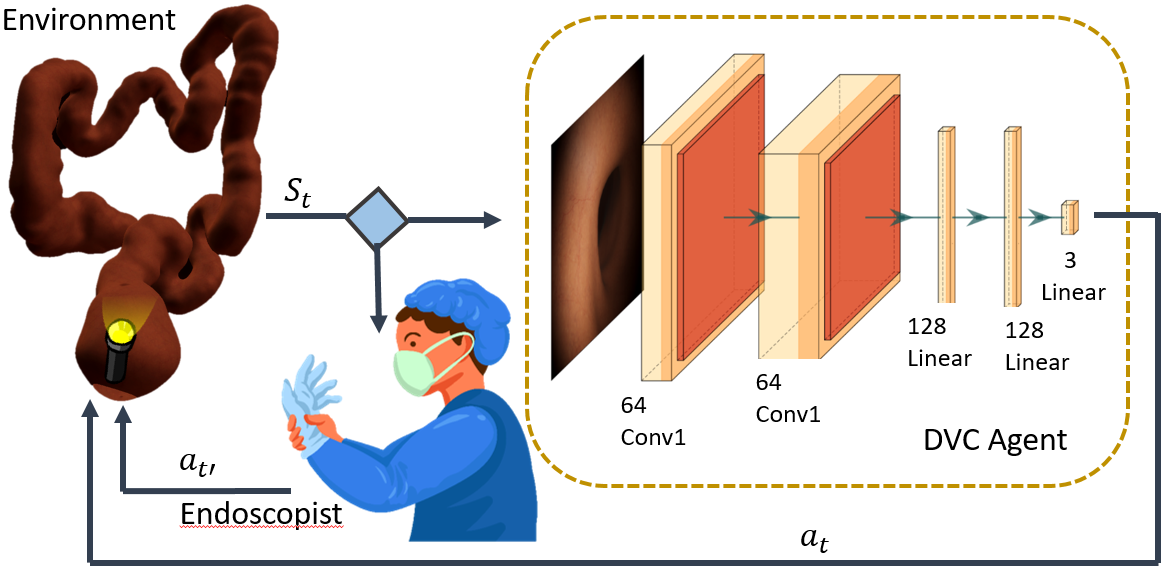}
	\caption{Deep Visuomotor Control (DVC) flow diagram. The environment provides a state observation $S_t$. The DVC agent uses the state input to generate an action $a_t$ that is applied to the environment. During the training phase, DVC learns a task-conditioned policy $\pi_{\phi}$ to perform autonomous colonoscopy navigation.
	In the evaluation phase, the clinicians can supervise and override DVC decisions through action $a_{t'}$.}
	\label{fig:title_figure} 
\end{figure}
Consequently, these procedures are more susceptible to human errors and increase the risks of tissue stretching and perforation, which are the prime cause of patient discomfort, and pain \cite{huang2021autonomous}. Moreover, some studies have reported work-related musculoskeletal injuries among endoscopists due to awkward neck and body posturing \cite{shergill2009ergonomics}. The shortage of adequately trained endoscopists with respect to the increasing clinical demand of colonoscopy procedures can lead to potential loss of human lives \cite{lee2014colonoscopy}.   

To overcome these limitations, wireless capsule endoscopes have been developed \cite{iddan2000wireless}; however, wireless devices lack the control of endoscopic point of view and increase the risk of missing some pathological areas \cite{ciuti2020frontiers}. Therefore, current research efforts are developing navigation systems using robotised FE, such as the STRAS system \cite{nageotte2020stras}, or Magnetic actuated FEs \cite{taddese2016nonholonomic}.
Robotised FE enables the introduction of automation technologies to enhance human operator abilities, in particular by adding autonomous navigation, which is the most time-consuming step of a routine colonoscopy procedure \cite{martin2020enabling}. 
This will allow the endoscopists to focus on the clinical aspect of the procedure rather than the manual control of FE, potentially improving the overall procedure outcome and reducing the training time \cite{huang2021autonomous}.

During the navigation phase, the clinician mainly uses visual feedback from the FE camera to advance through the lumen \cite{kragic2002survey}. 
A common gesture observed by endoscopists during a colonoscopy procedure is to centralise the target direction of the endoscope towards the lumen centre. Prior works on endoscopic navigation have built rule-based controllers to replicate this gesture by reducing the distance error between the image centre and the detected lumen centre \cite{stap2014image}.
These algorithms fail in situations when the tip of the endoscope approaches close to the colon wall. Such situations occur due to the highly deformable nature of the colon and the variable mobility introduced by patient movements, peristalsis and breathing, which lead to changes in lumen diameter and haustral folds due to which lumen detection is not trivial.
These situations require human interventions to find the correct motion direction, or they can be handled by adaptive exploration methods, as proposed in this work.

Originally postulated rule-based controllers are being progressively replaced by data-driven approaches such as Deep Reinforcement Learning (DRL), since they are able to provide some degree of adaptability \cite{richter2019open, tagliabue2020soft}.
However, the application of DRL in learning surgical task policies has been limited to low-dimensional physical state features such as robot kinematic data, which are widely accepted to be sample-efficient and trivial to learn \cite{tagliabue2020soft, tassa2018deepmind}. 
This paper proposed an image-based DRL approach for endoscopic control (Fig.~\ref{fig:title_figure}) focussing on learning the navigation task by devising an end-to-end policy to map the raw endoscopic images to the control signal of the endoscope, referred henceforth as Deep Visuomotor Control (DVC).
We primarily evaluate DVC control through a user study
with 20 expert GastroInstestinal (GI) endoscopists who perform the navigation task in a realistic virtual simulator.

The introduction of autonomous navigation can improve clinical practice, relieving clinicians from demanding cognitive and physical tasks. However,  in safety-critical areas, such as medical robotics, it is highly desirable to maintain human supervision to address ethical and legal concerns \cite{aiact}. Hence, it is essential to consider human-in-the-loop for DVC deployment in realistic surgical scenarios. Therefore, we conducted a second user study with 20 novice participants to demonstrate that non-expert users can easily supervise autonomous navigation, and DVC reduces the need for human intervention compared to a state-of-the-art method.



This work presents an initial study towards generating adaptive control for the colonoscopy procedure by proposing a DVC control policy for autonomous navigation and providing its performance evaluation with expert GI endoscopists.

The content of this paper is organised as follows: Sec.~\ref{sec:related_works} describes the related works, while implemented methods are explained in Sec.~\ref{sec: methods}. In Sec.~\ref{sec:experiments}, we elaborate on the experimental evaluation. Finally, the results and conclusions are discussed in Sec.~\ref{sec:results} and Sec.~\ref{sec:conclusions} respectively.
 
\section{RELATED WORKS} \label{sec:related_works}

The advantages of autonomous navigation in colonoscopy have prompted several studies in this field. In \cite{trovato2010development}, a screw-type colonic endoscope is developed, and motion adjustment is demonstrated using reinforcement learning. This study uses robot kinematics variables as state inputs; however, navigation through the straight segments was slow, and navigation through bends proved awkward due to the robot's size. 
Several studies have focused on magnetic guided endoscopes \cite{taddese2016nonholonomic, martin2020enabling, huang2021autonomous}.
In \cite{taddese2016nonholonomic}, navigation by following simple predefined trajectories is presented; hence extending this method to complex non-linear trajectories is challenging. 
Heuristic path planning algorithms are used in \cite{huang2021autonomous} to generate a feasible path in a colon map. This approach employs force-based real-time sensing to guide navigation. Force-based sensing is still not widely available in existing endoscopic devices; moreover, the interpretation of robotic actions without scene visualisation is challenging, hence not suitable for human supervision. 
In \cite{martin2020enabling}, a static perception model is developed, which extracts the centre of the lumen from raw image observation.
The control of endoscope position and orientation is imparted by a proportional controller that aligns the endoscopic image with the centre of the lumen. Similar rule-based controllers have been previously developed in \cite{stap2014image}; however, they require significant manual tasking for non-linear components such as analytically computing image jacobian, and interaction matrix \cite{saxena2017exploring}. Moreover, lumen detection could be unstable and prone to errors due to the dynamic nature of the colon and its sharp bends. Such scenarios require the vision-based control system to improve during policy training which is limited with hand-engineered features for perception \cite{saxena2017exploring}. Learning end-to-end visuomotor representations for direct control using DRL overcomes these limitations without separately designing perception and control models and offers the ability to improve model parameters while training \cite{ibarz2021train,levine2016end}.

Some studies have proposed frameworks for training DRL policies to automate surgical tasks \cite{richter2019open, tagliabue2020soft, xu2021surrol, scheikl2021cooperative} such as manipulation of rigid and deformable objects. 
These studies use simplified environments designed explicitly for robot-assisted surgery to learn the instrument control during the procedure. 
Recently, \cite{su2021multicamera} proposed a DRL method for optimising the endoscopic camera viewpoint.
These studies use low-dimensional state information for training DRL algorithms, such as kinematic values of the robot, the position of target etc. \cite{richter2019open, tagliabue2020soft, xu2021surrol, su2021multicamera}. 
In a real colonoscopy scenario, it is challenging to accurately capture the endoscope kinematics due to limits on the sensing capabilities \cite{huang2021autonomous}, and intra-operative guidance is solely based on visual feedback.

\begin{figure}[thpb]
	\centering
	\includegraphics[width=0.49\textwidth]{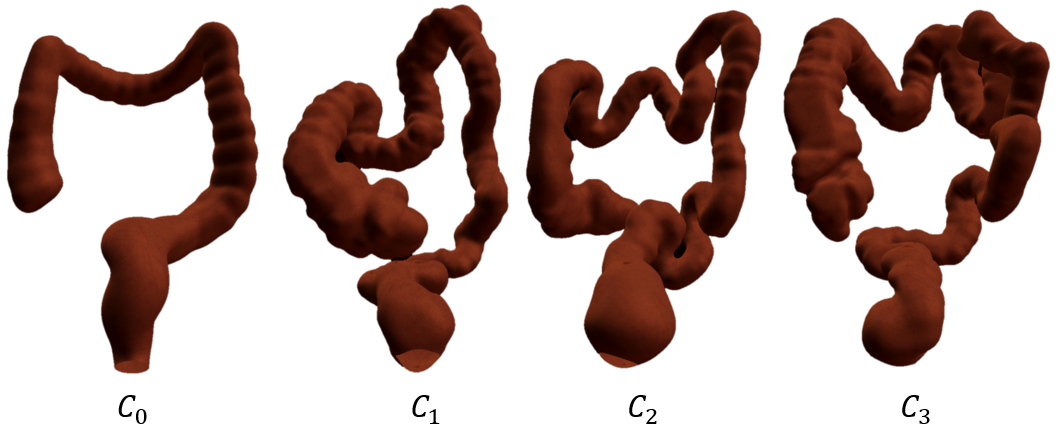}
    \caption{Colon models used in the experimental phase. (From left to right) ranked in increasing complexity order, $C_0$, $C_1$, $C_2$ and $C_3$ colon models. The model complexity is characterised by the centerline distance of the model from rectum to caecum, and the number of acute bending, i.e. $>$90 degree, which is estimated through visual inspection.}
	\label{fig:colon_models} 
\end{figure}

\section{METHODS} \label{sec: methods}
Our objective is to develop end-to-end joint training for perception and control to learn navigation policies that map raw endoscopic image observations directly to the control signals of the robotised FE (e.g. motor torques). We develop a realistic colonoscopy simulator with deformable tissue dynamics, described in Sec.~\ref{subsec: simulation}. Further, we explain the implementation details of DVC in Sec.~\ref{subsec:DVC}.

\subsection{Simulation platform} \label{subsec: simulation}
\textit{Colon simulation} - A public CT colonography dataset from the Cancer Imaging Archive \cite{clark2013cancer} is used to derive the colon models. 
The 3D models of the bowel are segmented in a semi-automated way \cite{jeong2021depth}. The segmented models are refined, and volumetrical and superficial meshes are generated. Realistic render textures are created using the real endoscopy images from KVASIR dataset \cite{incetan2021vr} and applied to the models. Thus, the reconstructed colon models are loaded into \textit{SOFA (Simulation Open Framework Architecture)}, where a realistic mechanical model based on real-time finite element analysis is generated, setting the simulation parameters in order to mimic the colon tissue behaviour \cite{christensen2015tensile}. Additionally, the collision detection between the endoscope and the colon is implemented, and physical constraints are included to realistically restrain the colon deformations. For creating high quality and realistic visual rendering, \textit{Unity3D} is adopted, exploiting the High Definition Render Pipeline (HDRP), which allows for additional visual effects such as reflection on the organ surface or vignetting the
peripheral darkening of the endoscopy image \cite{incetan2021vr}. Therefore, the \textit{SofaAPAPI-Unity3D} is used to interface the SOFA simulation with the Unity visual rendering.

\begin{figure}[thpb]
	\centering
	\includegraphics[width=0.40\textwidth]{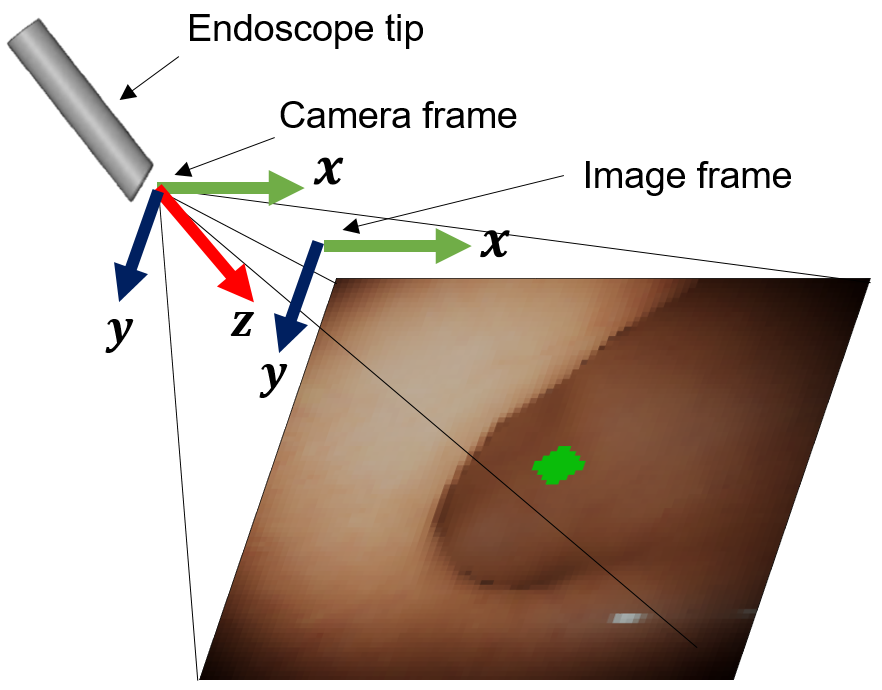}
	\caption{Representation of the local frame at the endoscope tip. The X-Y plane of the camera is parallel to the image frame, while the z-axis represents the direction of insertion. Tip bending is carried out on the X-Y plane while the roll is carried on the z-axis. DVC uses a low-resolution image as state input. The green region represents the detected lumen centre.}
	\label{fig:frame}
\end{figure}

\textit{Endoscope simulation} - 
We assume a scenario close to a magnetically guided FE where external magnets control the motion of the magnetic tip while the tether follows the tip passively \cite{taddese2016nonholonomic}. Hence, in this preliminary simulator version, we neglect the effect of the endoscope tether due to multiple collision points with the colon model that could lead to simulation instability. 
The endoscope tip is modelled as a rigid capsule with weight, length and diameter of 20g, 36mm and  14mm, respectively \cite{wang2021learning}. An angular drag of 4 rad/sec$^2$ is added to account for the frictional resistance \cite{wang2021learning}.
The endoscope tip embeds a camera and has four degrees of freedom for the motion as shown in Fig.~\ref{fig:frame}, i.e. translation (insertion/retraction), roll, bending in two perpendicular directions (pitch/yaw). 

\subsection{Deep Visuomotor control}  \label{subsec:DVC}

\textit{DRL background} - 
The colon navigation problem is formalised into a Markov Decision Process (MDP) represented by a tuple $(\mathcal{S}, \mathcal{A}, \mathcal{R}, \mathcal{P}, \gamma, T) $, where $\mathcal{S}$ denotes the state space, $\mathcal{A}$ is the action space, $\mathcal{P}$ is the transition probability distribution, $\mathcal{R}$ is the reward space, $\gamma \in [0, 1]$ is the discount factor and T is the time horizon per episode. At each timestep $t$, the environment produces a state observation $s_{t} \in \mathcal{S}$. The agent then generates an action $a_t \in \mathcal{A}$ according to a policy $a_t \sim \pi(s_t)$, and applies it to the environment to receive a reward $r_t \in \mathcal{R}$ \cite{sutton2018reinforcement}. As a consequence, the agent transitions to a new state $s_{t+1}$ sampled from the transition function $p(s_{t+1} | s_{t}, a_{t})$, $p \in \mathcal{P}$ or terminates the episode at state $s_{T}$. 

\begin{figure}[thpb]
	\centering
	\includegraphics[width=0.40\textwidth]{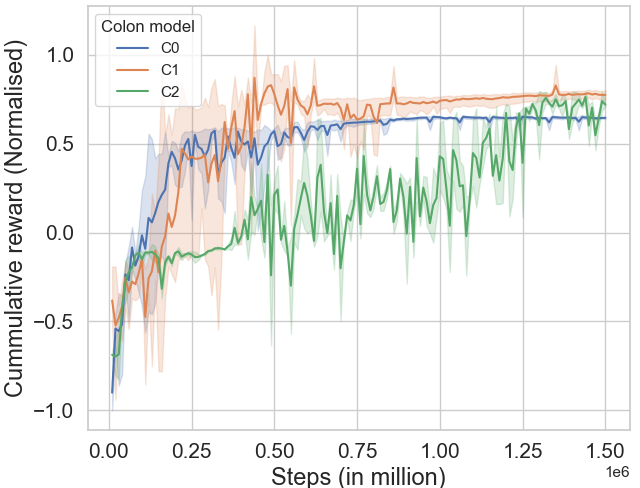}
	\caption{Learning curve of DVC trained on varying complexity of colon, using three colon models. Cumulative reward is normalised in the range $[-1,1]$. The shaded area spans the range of values obtained when training the agent starting from five different initialisation seeds.}
	\label{fig:learningcurve} 
\end{figure}

\textit{Learning algorithm} - The agent's goal is to learn a stochastic behaviour policy $\pi$ parametrised by $\phi$, $\pi_\phi: \mathcal{S} \rightarrow \mathcal{P}(\mathcal{A})$ to maximise the expected future discounted reward $E[\sum_{i=0}^{T-1} \gamma^{i} r_i]$. 
We chose PPO \cite{schulman2017proximal} as a consolidated DRL algorithm over Soft-Actor Critic \cite{haarnoja2018soft} and Deep deterministic policy gradient \cite{lillicrap2015continuous} due to overall returns in terms of wall-clock training time and hyper-parameter tuning. 
It is out of the scope of this work to propose a novel DRL method, while the main goal is to perform a user study to evaluate the performance of image-based DRL in colonoscopy navigation.
PPO consists of a value and a policy network that uses shared parameters to estimate the state value ($V$) and predict the action vector ($a$).
In the training session, the length of each episode is set as 10k iteration steps, $\gamma$ = 0.99, and the batch size and the learning rate hyperparameters are 64 and 3e-4, respectively. The PPO clip ratio was 0.2 with 4 mini-batches per epoch and 4 epochs per iteration. A loss term proportional to negative policy entropy was added, with a coefficient of 0.01. Each training lasted for 1.5 million iteration steps, which was the time taken for the reward function to converge (Fig.\ref{fig:learningcurve}).




\textit{Action space} -  The preliminary manual control of the endoscope revealed that if the endoscope is directed against the colon wall, especially at the sharp turns, the lumen is not visible. Hence, it is critical to avoid the translation of the endoscope in such scenarios. Therefore, we develop an action strategy, where a translation motion with a constant velocity of $v_{end} =$ 10mm/sec is carried out only when the lumen is detected. The action space consists of discrete angular rotation values in the three degrees of freedom at the endoscope tip, $\delta \theta_j =  \alpha$, $\alpha \in \{0, -1, +1\}$ in the $j^{th}$ spatial dimension.
In the tip local reference frame, $j \in {x,y,z}$ corresponds to the orientation alignment in the horizontal and vertical directions in the image plane and the endoscope roll, respectively (Fig.~\ref{fig:frame}).
In cases when the lumen is not visible, the translation velocity of the endoscope is set to zero, and the agent carries out orientation changes to detect the lumen. 

\textit{Observation space and policy} - The sensory input to the DVC agent is composed of a downscaled endoscopic image.
The RGB images rendered by the endoscopic camera (1024x1024 pixels) is downscaled to 128x128 pixels to reduce the computational complexity of the training DVC network. 
The policy $\pi_\phi$ is represented by a CNN architecture, consisting of two convolutional layers (Fig.~\ref{fig:title_figure}) for encoding visual scene representations.
The network details are publicly available on the project website\footnote{https://github.com/Ameyapores/DVC}. 
The output of the convolutional layers  
are fed into a combination of fully connected layers and Long Short-Term Memory (LSTM) layer to represent time-dependent behaviour, each with 128 rectified units, followed by the linear connections to the output logits $\pi_t$ for each action $a_{t}$ and values estimate $V_t$. A softmax function transforms the logits to action probabilities. The complete network is trained end-to-end to acquire task-specific visual features.
\begin{figure}[thpb]
	\centering
	\includegraphics[width=0.49\textwidth]{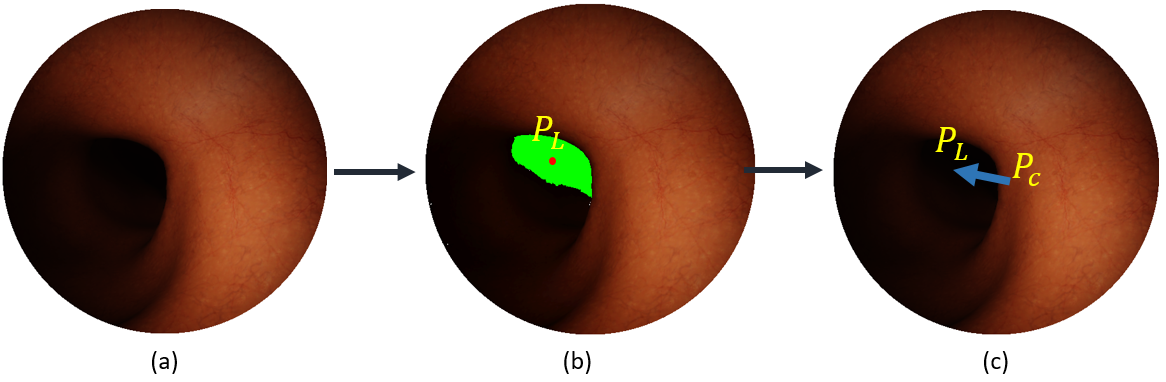}
	\caption{Proposed adaptive threshold segmentation pipeline for lumen detection. Each RGB frame captured by the endoscopic camera is passed through the adaptive filter to detect the dark pixels a) original RGB frame b) Image mask for the detected lumen (in green) c) distance vector between the image centre $P_c$ and the centroid of the detected darkest regions $P_L$.}
	\label{fig:image_segmentation} 
\end{figure}

\begin{figure}[thpb]
	\centering
	\includegraphics[width=0.40\textwidth]{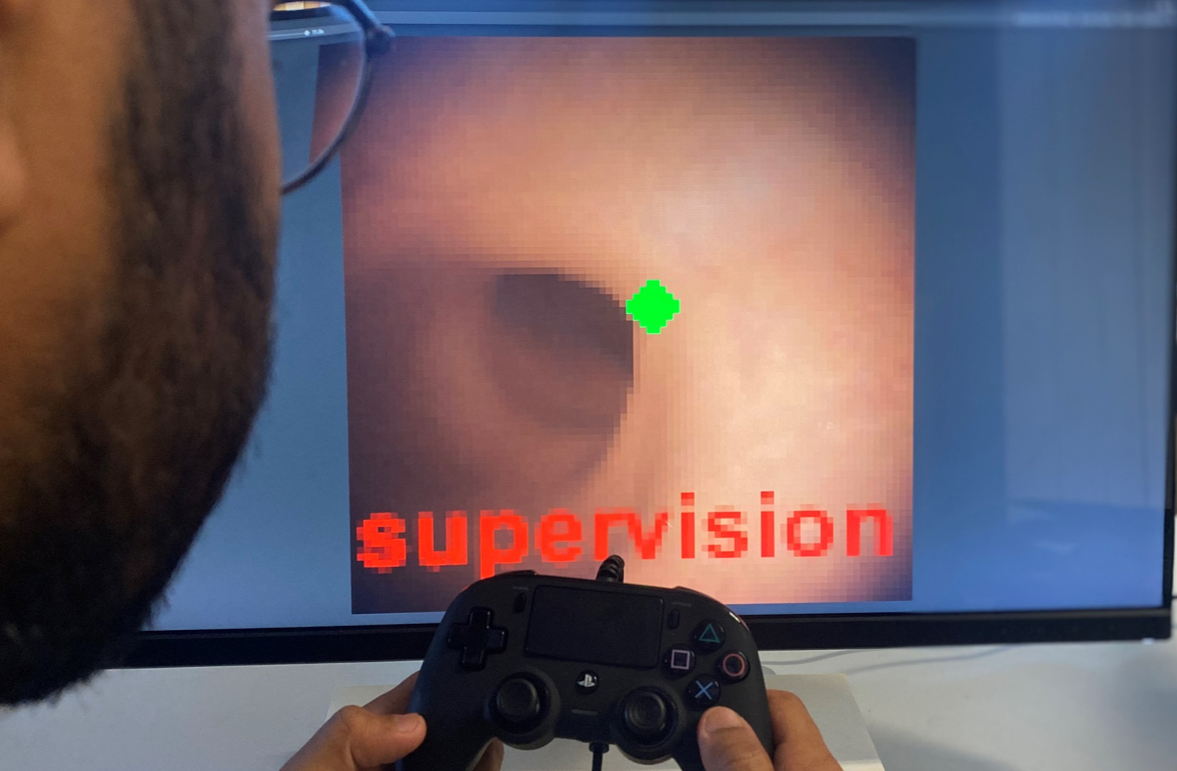}
    \caption{Manual supervision through a joystick while navigating by autonomous control strategies. \textit{supervision} is printed on the screen, indicating the switch to manual control. When the endoscope is oriented towards the lumen (green point), the user can give back the control to the autonomous agent. A low-resolution (128x128 pixels) image is displayed to facilitate interpretability of machine decisions, however users have the option to change to high resolution (1024x1024 pixel) display.}
	\label{fig:supervision} 
\end{figure}

\begin{footnotesize}
\begin{table*}
\centering
\vspace{15pt}
\caption{Navigation parameters used for validation with their description} \label{tab:autonomy_table} 
\begin{tabular}{p{0.3cm}p{1.8cm}p{14cm}}
\noalign{\smallskip }\hline\noalign{\smallskip}
 & \makecell{Navigation\\metrics} & Description \\
\toprule
1 & Time of insertion (TOI) & TOI is measured from the time point where the initial movement of the endoscope is detected to the time point when the caecum is reached. \\
\midrule
2 & Perforation & Perforation refers to the scenario when excessive force is applied on the colon wall (especially at the turning point) that can lead to severe injuries. 
Studies based on tensile property analysis of human rectal tissue reported the maximum elongation of 62\% \cite{christensen2015tensile}. The average diameter of the colon models used is 5cm, hence a threshold of $\delta d = 3cm$ is decided to classify the deformation as perforation. \\
\midrule
3 & Normalised distance travelled & Distance travelled is crucial as multiple backward motions, reversing the direction, can lead to suboptimal trajectories.
The distance travelled is measured using the position values of the endoscope tip. This distance is normalised by the centerline distance of the colon model in order to compare among different colon models. Normalised distance above 1 indicates a path distance longer than the centerline path, while a normalised distance below 1 indicates a shorter path than the centerline was followed. \\
\midrule
4 & Average LD & Lumen centralisation is believed to create smooth insertion trajectories hence the lumen distance in the image plane is recorded at each timepoint. This distance is normalised by the size of the image to get a value in [0,1]. Lumen distance value 0 denotes image centre ($P_c$) coincides with the detected lumen ($P_L$), and value 1 denotes that the detected lumen is at the farthest point. \\
\midrule

\end{tabular}
\end{table*}
\end{footnotesize}

\textit{Reward function} - 
The goal of navigation is to reach the end of the colon without any significant complications.
Visuomotor control should be able to track the colon during the whole procedure. Successful tracking requires the lumen centre $P_L$ to be close to the image centre $P_c$.
Hence, a dense reward $r_t(s_t,a_t)$ is designed as follows:
\begin{equation}
r_t(s_t,a_t) = \begin{cases}
C(1-(||P_L - P_c||_2/D_{max})), & L = 1\\
-1, &           L=0
\end{cases}
\end{equation}
where $D_{max} = 1/2*(Image width) = 64$, is the normalisation factor which is the maximum distance possible, $L$ represents the lumen detection flag, (1 denotes lumen detected, 0 denotes no lumen detected), the hyperparameter $C$ is chosen as 1. Moreover, the agent is awarded a reward of +10 if the colon end is reached and -10 if it returns to the original starting point, to encourage the agent to move unidirectional towards the caecum.
To detect the colon lumen in the endoscope image, we build a threshold segmentation algorithm that runs in real-time at 30fps based on \cite{wang2014lumen}, where the image is first segmented to detect the darkest and most distinct region, with the presumption that this area contains the distal lumen with high probability (Fig.\ref{fig:image_segmentation}). The segmentation is performed by converting the RGB image to greyscale and cropping a circular region centred with the image and a diameter equal to the image width to remove the vignette effect on the corner. 


\begin{figure*}[thpb]
	\centering
	\subfloat[Lumen distance]{
    \centering
    \includegraphics[width=0.24\linewidth]{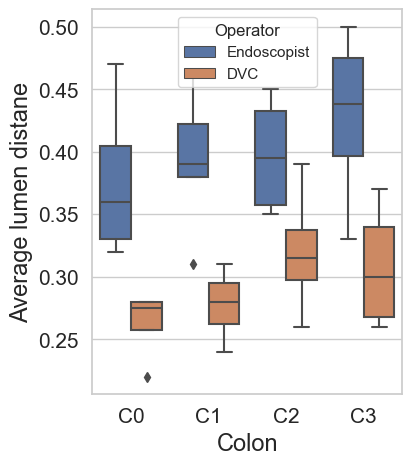}}
    \subfloat[Perforations]{
    \centering
    \includegraphics[width=0.24\linewidth]{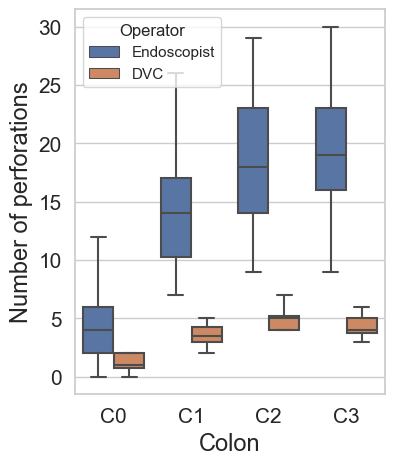}}
    \subfloat[Normalised distance]{
    \centering
    \includegraphics[width=0.24\linewidth]{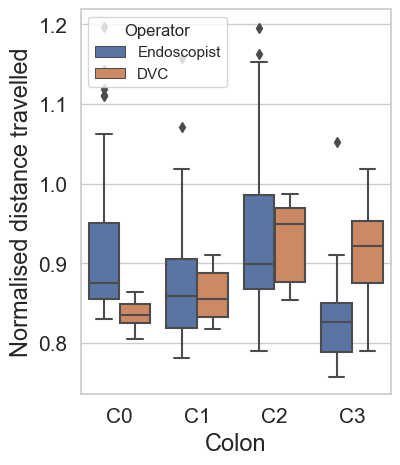}}
    \subfloat[Time of insertion]{
    \centering
    \includegraphics[width=0.24\linewidth]{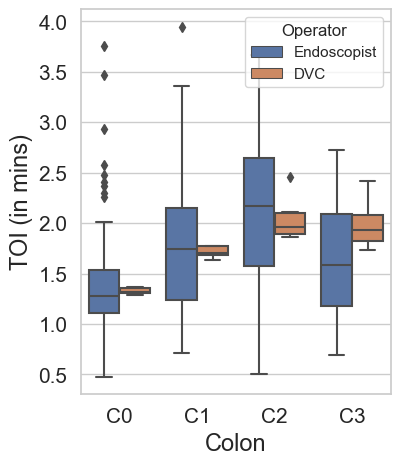}}
	\caption{Navigation performance comparison plots between DVC and endoscopists. Several parameters are plotted a) Lumen distance, b) perforations, c) Normalised distance, d) Time of insertion.}
	\label{fig:compare_plots} 
\end{figure*}

\section{EXPERIMENTAL EVALUATION} \label{sec:experiments}
The experimental goal is to compare the navigation performance of the DVC agents, the baseline method of rule-based control \cite{martin2020enabling} and the endoscopists. Hence, we create a pipeline where the position and orientation values of the endoscope, lumen distance in the image space, colon deformations, and camera image can be recorded within our developed simulator. During the experiments, these parameters were synchronised and recorded for future processing using \textit{labstreaminglayer} software, which is a unified system for the collection of time-series measurements \cite{merino2020easy}. 

\textit{Endoscopist data acquisition} -  
A group of 20 expert GI endoscopists (more than four years of experience) were asked to make navigation attempts in the colonoscopy simulation scene developed in Sec.~\ref{subsec: simulation}. 
Due to the time constraints and COVID regulations at the hospital\footnote{Ospedale Le Molinette (Torino, Italy)}, four colon models were selected considering the opinion of domain experts to represent progressively more complex scenarios (Fig.~\ref{fig:colon_models}).
Each endoscopist was instructed to navigate the colon models from the rectum to the caecum using a through PlayStation (Sony Interactive Entertainment, USA) joystick device. The colon model $C_0$, which depicts a simplified colon model that conforms with the shape and size of the average human colon, was used to familiarise the endoscopists with the controls before initialising the trials. 
The trials started with endoscopist attempts on the $C_1$ colon, followed by randomised attempts on $C_2$ and $C_3$ colon. The randomness between $C_2$ and $C_3$ colon was introduced to identify performance bias based on the colon model.

\textit{Training DVC agents} -  
We conduct three experiments to validate the DVC. First, our aim is to determine the sample efficiency of training on different levels of colon complexity. Hence, we train DVC agents separately using the same models employed during the endoscopist experiment.
Second, to establish a comparative analysis between the DVC and endoscopists, a similar experimental workflow was followed as in the endoscopist experiments, where the DVC was only trained on $C_0$ ($DVC_{C_{0}}$) and tested on $C_1$, $C_2$ and $C_3$ colons.
Third, the DVC was trained on the $C_0$ model followed by training on the $C_1$ ($DVC_{C_{0}+C_{1}}$)  to test if training on a complex colon after a simple one improves performance. To keep the overall iteration steps for DVC training at 1.5 million, the training on $C_0$ was terminated after 1 million iteration steps and loaded back to train on $C_1$ for 500k iteration steps (Table.~\ref{Tab:abb}).
\begin{table*}
\centering
\vspace{15pt}
\caption{Comparison between DVC and Endoscopists}
\label{Tab:abb}
\begin{tabular}{ c c c c c c c c c c c c }
\hline\noalign{\smallskip}
      & \multicolumn{4}{c}{$DVC_{C_0}$} && \multicolumn{4}{c}{$DVC_{C_{0}+C_{1}}$} \\
\noalign{\smallskip }\cline{2-5} \cline{7-10} \noalign{\smallskip} 
      & \makecell{Average\\LD} & Perforation & TOI & \makecell{Normalised \\ distance} && \makecell{Average\\LD} & Perforation & TOI & \makecell{Normalised \\ distance} \\
\noalign{\smallskip}\hline\noalign{\smallskip}
$C_0$ & 0.27$\pm$0.01 & 0.5$\pm$0.25 & 1.37$\pm$0.05 & 0.84$\pm$0.02 && 0.24$\pm$0.02 & 1$\pm$1 & 1.32$\pm$0.03 & 0.84$\pm$0.02 \\
$C_1$ & 0.30$\pm$0.01 & 3.3$\pm1.5$ & 1.74$\pm$0.04 & 0.88$\pm$0.08 && 0.25$\pm$0.01 &  3.3$\pm$0.5 & 1.70$\pm$0.07 & 0.82$\pm$0.03  \\
$C_2$ & 0.36$\pm$0.03 & 5$\pm$1 & 2.22$\pm$0.2 & 0.97$\pm$0.03 && 0.28$\pm$0.01 & 4.6$\pm$0.5 & 1.89$\pm$0.03 & 0.85$\pm$0.01 \\
$C_3$ & 0.35$\pm$0.02 & 4.6$\pm$0.5 & 2.15$\pm$0.23 & 0.92$\pm$0.08 && 0.29$\pm$0.03 & 3$\pm$1 & 1.78$\pm$0.05 & 0.89$\pm$0.09  \\
\hline
Mean & 0.31$\pm$0.04 & 5.0$\pm$1.2 & 2.20$\pm$0.75 & 0.90$\pm$0.04 && \textbf{0.23$\pm$0.04} & \textbf{4.3$\pm$1.2} & \textbf{1.96$\pm$0.59} & \textbf{0.86$\pm$0.04} \\
\noalign{\smallskip}\hline
\end{tabular}
\end{table*}

\textit{Supervision} - 
20 novice participants (no endoscopy experience) were asked to supervise the performance of the rule-based controller agent and the DVC agent. The experimental workflow consisted of three trials; inside each trial, the participants attempted to navigate $C_1$, $C_2$ and $C_3$ colon models. Each trial was characterised by one of the following control strategies. 
\begin{enumerate}
  \item Manual control: Participants were instructed to exclusively control the endoscope using a joystick during the entire duration of the procedure.
  \item Rule-based baseline \cite{martin2020enabling}: A proportional controller is generated for orientation control that aligns the image center ($P_c$) to the detected lumen ($P_L$) as follows:
  \begin{equation}
  \delta \theta = \beta \begin{bmatrix}
P_{L_x} - P_{c_x}\\
P_{L_y} - P_{c_y}
\end{bmatrix}
  \end{equation}
  We refer to the distance between $P_L$ and $P_c$ as Lumen Distance (LD).
  \item DVC: A fully trained $DVC_{C_0}$ was deployed.
\end{enumerate}
In control strategy 2, the rule-based controller indicates the requirement of manual supervision when the lumen centre is not detected (Fig.~\ref{fig:supervision}). 
In control strategy 3, the agent is given ($\Delta_t = 50$) iteration steps to search for lumen when the lumen is not detected. After $\Delta_t$ steps, the DVC notifies the requirement of human supervision, and manual control is activated. In both the control strategies, the user has an override option to take control when unsafe behaviour is encountered, e.g. collision with the colon wall or direction of motion reversed. 
Once the manual control is active, the participants can navigate the endoscope safely and give back the control to the DVC or rule-based controller.
During each attempt, the number of interventions by the participant was recorded. After all the trials, users were asked to complete a NASA Task Load Index (TLX) questionnaire \cite{hart1988development}, to score human-perceived workload.


\begin{table}
\centering
\caption{NASA Task Load Index for novice users. Lower score indicate good user experience}
\label{Tab:task}
\begin{tabular}{ c c c c }
\noalign{\smallskip }\hline\noalign{\smallskip}
 & \makecell{Manual\\control} & \makecell{Rule-based\\control} & DVC \\
\noalign{\smallskip}\hline\noalign{\smallskip}
Mental demand & 63 & 33 & \textbf{18}\\
Physical demand & 65 & 38 & \textbf{9} \\
Temporal demand & 30 & 47 & \textbf{17} \\
Performance & 25 & 34 & \textbf{12}  \\
Effort & 57 & 38 & \textbf{10}  \\
Frustration & 42 & 41 & \textbf{12} \\
Mean workload & 47 & 38 & \textbf{13} \\
\noalign{\smallskip}\hline
\end{tabular}
\end{table}
\begin{figure*}[thpb]
	\centering
    \subfloat[$C_1$]{
    \centering
    \includegraphics[width=0.30\linewidth]{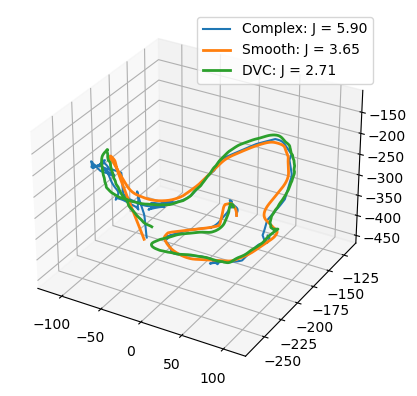}}
    \subfloat[$C_2$]{
    \centering
    \includegraphics[width=0.30\linewidth]{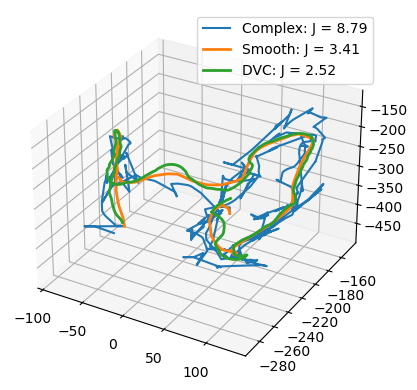}}
    \subfloat[$C_3$]{
    \centering
    \includegraphics[width=0.30\linewidth]{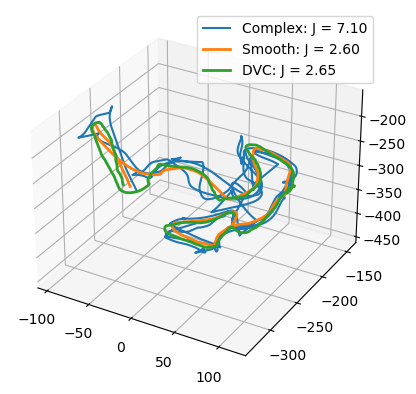}}
	\caption{Trajectory plot of DVC, complex and smoothest endoscopist performance for a) $C_1$ b) $C_2$ 3) $C_3$ models respectively.}
	\label{fig:trajectory_plots}
\end{figure*}

\textit{Data Analysis} - Four different parameters are used to score the navigation performance. 
Time of Insertion (TOI) and the number of colon perforations are qualitative assessment measures for colonoscopy procedures \cite{kaminski2017performance}, while average LD and the normalised distance travelled are the two metrics devised in this study to measure the accuracy of the trajectories. The details of each parameter are elaborated in Table.~\ref{tab:autonomy_table}. When the user or DVC reversed its direction of motion and returned to the rectum or perforated heavily to destabilise the colon model, it was considered a failed navigation attempt.

\section{RESULTS AND DISCUSSION} \label{sec:results}

The learning curves when DVC is trained on different levels of colon complexity are presented in Fig.~\ref{fig:learningcurve}. $C_0$ represents a simplistic model; hence, the DVC agent reaches high reward values in relatively fewer steps than in other colon models. 
A high reward indicates that the agent successfully learns to complete the navigation task.
Whereas $C_2$ represents high complexity, the agent requires 1.2 million steps for high-reward convergence. The $C_1$ training curve lies between $C_0$ and $C_2$. This suggests that the training time is related to colon complexity.
However, note that $DVC_{C_0}$ can navigate other complex colon models, i.e. it acquires task-specific features that can generalise to other colon models (Table.~\ref{Tab:abb}). 

\textit{Comparative analysis} - The performance data of 20 endoscopists was acquired while 10 different DVC agents were trained on the $C_0$ starting with a different random seed. The realism of the simulation was confirmed and validated by expert clinicians. Additionally, all the users positively evaluated the joystick used to navigate the endoscope inside the colon as intuitive, user-friendly and easy to learn. Fig.~\ref{fig:compare_plots} shows the comparison of the average LD, the number of perforations, the completion time and the normalised distance travelled.
There is a significant difference in the average LD and the number of perforations between the endoscopists and the DVC. DVC shows precise tip centralisation and less number of perforations compared to endoscopists. One of the reasons for this difference is that clinicians tend to push the colon wall at acute bends of colon junctions (See supplementary video). This is a gesture sometimes clinicians follow due to the rigid constraints of the clinically available FEs. 
Whereas DVC is trained on reward feedback to minimise LD, it stays centralised to avoid contact with the wall. For the normalised distance and TOI, a substantial difference is not noted. There is more variance observed in the performance of the endoscopists. Some endoscopists followed a convoluted trajectory that increased the normalised distance and time of insertion, while others followed smoother trajectories that resulted in the lower normalised distance and TOI. Fig.~\ref{fig:trajectory_plots} shows the most complex and smoothest trajectories demonstrated by the endoscopists and the trajectory executed by $DVC_{C_0}$ for the $C_1$, $C_2$ and $C_3$ colons.
The smoothness of a trajectory is estimated using a jerk index J ($cm/sec^3$) which characterises the average rate of change of acceleration in a movement \cite{shadmehr2005supplementary}.
Human operators tend to show wide variance in performing optimal trajectories, while DVC performance stays in the average range.

The result of splitting the training into two colon models $DVC_{C_0+C_1}$ and evaluating on other colon models are shown in Table.~\ref{Tab:abb}. 
There is an improvement in the lumen detection performance for $DVC_{C_0+C_1}$ in comparison to $DVC_{C_0}$. $DVC_{C_0}$ reaches high rewards at 500k iteration steps; hence there is no additional feedback to improve the performance. We speculate that the agent reaches suboptimal local minima, while,
when the DVC trained on $C_0$ is loaded to train on $C_1$, it encounters acute bends, which offers the potential to maximise the cumulative reward. There is no considerable improvement on other navigation parameters, i.e. perforation, TOI and normalised distance.

\textit{Supervision} - The human interventions are divided into two parts. First, where the user overrides the control due to unsafe behaviour and second, where the system demands human supervision. 
The average human intervention required for rule-based baseline was $5\pm1.8$ for human override and $2.5\pm1.5$ when the system demanded human control, while for DVC, the average number of human interventions are $0.1\pm0.5$ for human override and $0.05\pm0.2$ when the system demanded human control. 
This difference is attributed to DVC's adaptability to search for new insertion directions when the lumen is not easily detected, whereas the rule-based controller lacks this ability.
The NASA-TLX for each control strategy is shown in Table.~\ref{Tab:task}. Regarding ease of use, participants found manual control and rule-based controller more demanding in all task load categories, while a substantial workload reduction is observed for DVC. 


\section{CONCLUSIONS}  \label{sec:conclusions}

Prior works on autonomous colonoscopy navigation use heuristic control policies that fail to adapt to situations where detecting lumen is not straightforward and requires frequent human intervention. This article proposes an end-to-end DVC that learns a mapping between the endoscopic images and the endoscope's control signal, such as tip orientation. DRL has been applied in the surgical domain; however, these works use low-dimensional physical robotic state features that are challenging to obtain using FEs.
To evaluate the navigation performance of DVC, motion data from 20 GI endoscopists was acquired and compared to DVC control. Our experimental validation shows an equivalent performance in terms of the time of insertion and the distance travelled. However, DVC reduces the number of perforations and shows efficient lumen tracking, improving safety. Furthermore, we conducted a second novice user study to demonstrate that supervision of DVC control significantly reduces the user workload with overall performance comparable to expert GI endoscopists.


On the contrary, there are some limitations of this work. First, it is not straightforward to know the direction of motion of the endoscope. Hence, our newer version of the virtual scene will simulate the endoscope body dynamics, providing the insertion length.
Second, 
if the robot needs to learn from raw image observations, it also needs to evaluate the
reward function from raw image observations, which itself requires a hand-designed perception system. This can be mitigated by using online user interaction through human-in-the-loop reinforcement learning \cite{chen2022asha}. 
The results obtained in Table.~\ref{Tab:abb} presents an opportunity to study curriculum learning-based setup \cite{bengio2009curriculum}, where colon navigation can be trained in increasing levels of colon complexity. 
Our future work will demonstrate the formal validation of the realism of the proposed virtual simulator.








\bibliographystyle{IEEEtran}{

\bibliography{root}}


\end{document}